\documentclass[letterpaper]{article} 
\usepackage{aaai2027}  
\nocopyright
\usepackage[hyphens]{url}  
\usepackage{graphicx} 
\usepackage{natbib}  
\usepackage{caption} 
\usepackage{algorithm}
\usepackage{algorithmic}
\usepackage{amsmath} 
\usepackage{booktabs}
\usepackage{enumitem}
\usepackage{newfloat}
\usepackage{listings}
\DeclareCaptionStyle{ruled}{labelfont=normalfont,labelsep=colon,strut=off} 
\floatstyle{ruled}
\newfloat{listing}{tb}{lst}{}
\floatname{listing}{Listing}

\usepackage{booktabs}

\title{\textsc{ConsultMind}: Towards Automated Diagnostic Consultation via Uncertainty-Aware Reasoning}

\author{
Xiao Sun\textsuperscript{\rm 1},
Yuming Yang\textsuperscript{\rm 1},
Yun Chen\textsuperscript{\rm 2},
Jiang Zhong\textsuperscript{\rm 1,$\dagger$},
Junnan Zhu\textsuperscript{\rm 4},
Xinyi Jiang\textsuperscript{\rm 6},
Haoyang Zeng\textsuperscript{\rm 1},
Ruirui Chen\textsuperscript{\rm 7},
Yining Wang\textsuperscript{\rm 3},
Xinyu Zhou\textsuperscript{\rm 5},
Rong Tang\textsuperscript{\rm 8},
Kaiwen Wei\textsuperscript{\rm 1,$\dagger$}
}

\affiliations{
\textsuperscript{\rm 1}College of Computer Science,
Chongqing University\\
\textsuperscript{\rm 2}College of Computer Science,
Hunan University
\quad
\textsuperscript{\rm 3}AI Labs, Unisound\\
\textsuperscript{\rm 4}Institute of Automation,
Chinese Academy of Sciences
\quad
\textsuperscript{\rm 5}Psycharity,
Chongqing Medical University\\
\textsuperscript{\rm 6}School of Computer Science and Engineering,
University of New South Wales\\
\textsuperscript{\rm 7}Institute of Advanced Intelligence and Computing
(IAIC),\\
Agency for Science, Technology and Research (A*STAR), Singapore\\
\textsuperscript{\rm 8}Technological Information,
Chongqing Center for Disease Control and Prevention\\[3pt]
sunx@stu.cqu.edu.cn,
\{zhongjiang, weikaiwen\}@cqu.edu.cn
}

\begin{document}

\maketitle

\begingroup
\renewcommand{\thefootnote}{$\dagger$}
\footnotetext{Co-corresponding authors.}
\endgroup

\begin{abstract}
Diagnostic consultation is an online sequential decision-making process in which clinicians gather evidence through patient interaction until a diagnosis is sufficiently supported. Automating this process requires adaptive inquiry and interpretable decisions. Bayesian networks offer a natural foundation by updating diagnostic posteriors as evidence accumulates, but their use in open-ended consultation raises two challenges: linking diagnostic hypotheses to potential inquiries and translating evolving posteriors into consultation decisions. We introduce \textsc{\textbf{AutoDisym}}, an automated pipeline that integrates diagnostic knowledge with heterogeneous diagnosis-labeled clinical narratives to construct a Disorder--Symptom Bayesian Network (DSBN). Building on the DSBN, we propose \textsc{\textbf{ConsultMind}}, an uncertainty-aware framework that updates disorder posteriors after each response and uses posterior uncertainty to guide inquiry and diagnosis. We evaluate both methods across psychiatry, respiratory medicine, fever clinics, and three public datasets. The results show that {AutoDisym} can automatically construct high-quality DSBNs and that {ConsultMind} consistently improves diagnostic performance and explanation soundness. For example, {AutoDisym} achieves macro-averaged F1 scores of 81.37 for canonical symptoms and 72.19 for manifestations using {GPT-5.6-Sol}. {ConsultMind} improves Top-1 and Top-3 diagnostic accuracy by up to 22.15 and 37.89 percentage points, respectively. Physician evaluation further shows that {ConsultMind} improves the quality of ranking explanations, differential diagnoses, and diagnosis rationales across LLMs of different scales. This work offers a promising approach to automatic diagnostic consultation.
\end{abstract}


\section{Introduction}

\begin{figure}[t!]
    \centering
    \includegraphics[width=1\linewidth]{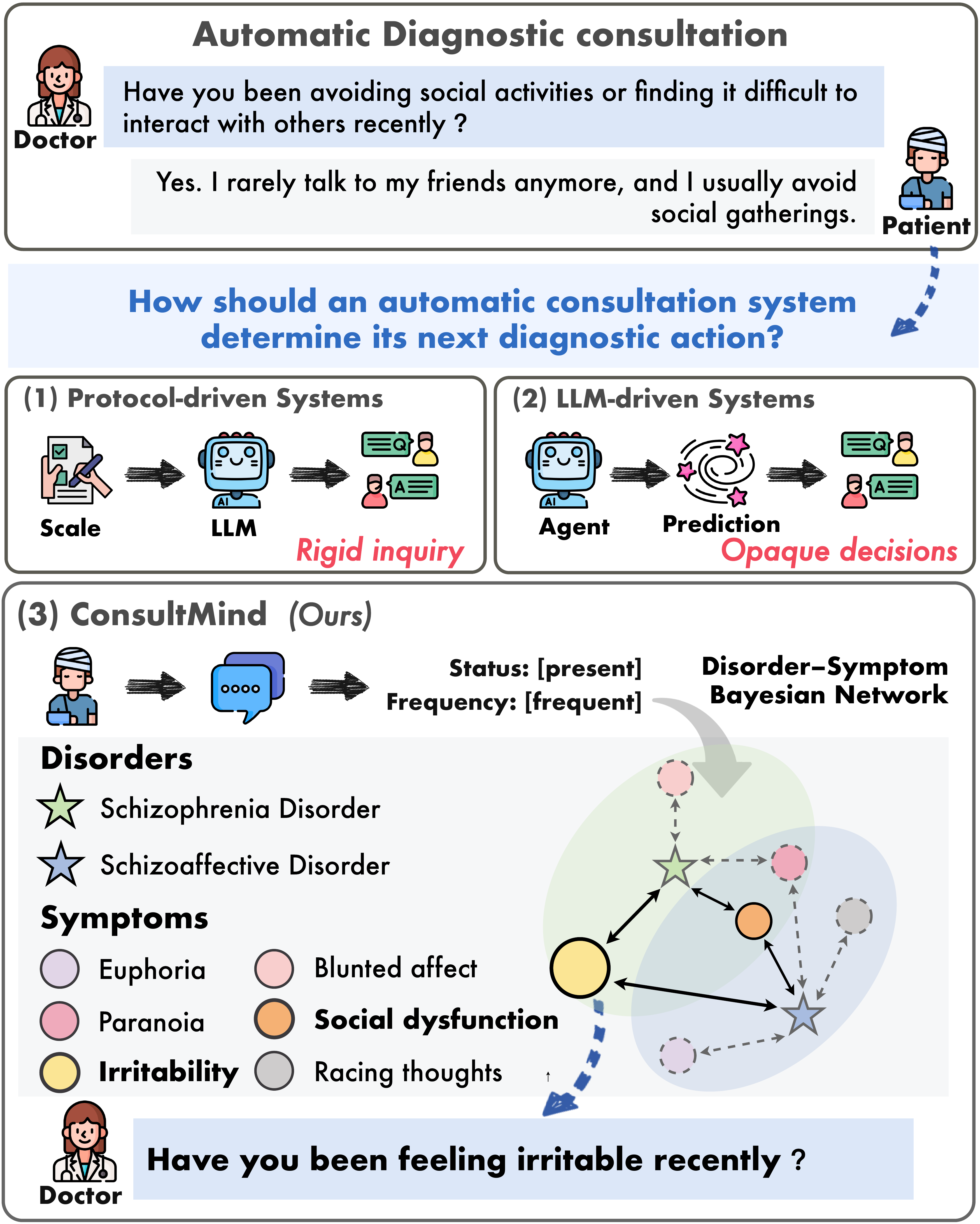}
\caption{\textbf{Comparison of automated diagnostic consultation methods.} Protocol- and LLM-driven systems suffer from rigid inquiry and opaque decisions, respectively, whereas \textsc{ConsultMind} enables adaptive and interpretable consultation through uncertainty-aware reasoning.}
    \label{fig:intro}
    \vspace{-16pt}
\end{figure}

Diagnostic consultation often begins with patients providing partial or imprecise
descriptions of their conditions, requiring clinically relevant information to
be recovered through multi-turn interaction~\citep{zeng2020meddialog,shi2023midmed}.
Clinicians respond with targeted questions about symptoms and medical history,
progressively narrowing plausible diagnoses~\citep{tang2016inquire,wei2018task,yuan2024efficient}.
Consultation is therefore an iterative decision-making process that gathers
evidence until it is sufficient to support a diagnosis~\citep{li2024mediq,werthaim2026benchmark}.

Recent advances in large language models (LLMs) have accelerated research on automatic diagnostic consultation~\citep{tu2024towards,saab2026advancing}, in which a system interacts with a patient through successive inquiries until it can formulate a diagnostic hypothesis~\citep{li2024mediq,werthaim2026benchmark}. Existing systems generally follow either a protocol-driven~\citep{wei2018task} or an LLM-driven paradigm~\citep{ren2025diallms}. Protocol-driven systems organize consultations around structured symptom spaces, symptom checklists, or diagnostic pathways, offering standardized and controllable procedures~\citep{yuan2024efficient}. LLM-driven systems generate questions and diagnostic outputs from the dialogue context, allowing greater adaptation to patient-specific information~\citep{sanghvi2026medxagent}.

Despite these strengths, recent evaluations have identified persistent weaknesses
in the inquiry quality, clinical reasoning, and decision reliability of LLM-based
consultation systems~\citep{johri2024craft,johri2025evaluation}.
Existing systems have two key limitations. (1) \textit{Rigid inquiry}.
Protocol-driven systems confine inquiries to predefined sequences or transition
rules~\citep{xia2020generative}, potentially overlooking
diagnostically informative symptoms that fall outside the prescribed paths.
(2) \textit{Opaque decisions}. LLM-driven systems derive consultation decisions
directly from the dialogue context, leaving the evolving diagnostic state and
the rationale for each decision opaque~\citep{gong2025dialogue,qiao2026medconsultbench}.
Diagnostic consultation, however, is an online sequential decision-making
process in which clinicians update their diagnostic assessment after each
patient response and use it to guide the next decision~\citep{sun2026mentalseek}.
Existing systems therefore lack a mechanism that tracks the evolving diagnostic
state, adapts inquiry accordingly, and makes each decision interpretable.

As shown in Figure~\ref{fig:intro}, within a structured variable space, Bayesian inference updates disorder posteriors as evidence accumulates, while explicit probabilistic dependencies make the basis of each update inspectable. Diagnostic consultation, however, lacks a predefined variable space because patient descriptions and potential inquiries are open-ended. Applying Bayesian networks to this setting therefore raises two unresolved questions: \textit{(1) how to link diagnostic hypotheses to potential inquiries} and \textit{(2) how to translate evolving posteriors into consultation decisions}.

To address the first question, we introduce \textsc{\textbf{AutoDisym}}, an automated pipeline that integrates diagnostic knowledge with diagnosis-labeled clinical narratives to construct a Disorder--Symptom Bayesian Network (DSBN). The DSBN represents diagnostic hypotheses as candidate disorders and potential inquiries as clinically grounded symptom variables. Through collaboration among specialized agents, {AutoDisym} derives a canonical symptom schema from diagnostic knowledge and maps symptom observations from heterogeneous clinical narratives, including consultation dialogues, electronic medical records, and clinical case, onto this schema. Finally, {AutoDisym} refines the schema through corpus-level feedback from recurrent unmatched observations and uses the grounded observations to parameterize the DSBN.

Building on this, we further propose \textsc{\textbf{ConsultMind}}, an uncertainty-aware reasoning framework that explicitly grounds each consultation decision in the evolving disorder posterior. {ConsultMind} comprises two mechanisms: (1) \textit{Uncertainty Awareness}, which updates this posterior after each patient response and quantifies the remaining diagnostic uncertainty; and (2) \textit{Adaptive Reasoning}, which supports three inquiry strategies and dynamically selects among them based on the evolving posterior state. \textit{Exploration} seeks potentially relevant symptoms; \textit{Differentiation} targets symptoms that distinguish competing hypotheses; and \textit{Consolidation} gathers further evidence for the leading hypothesis. Together, these mechanisms enable {ConsultMind} to guide an LLM through successive consultation decisions and provide a diagnosis once sufficient evidence has accumulated.

We evaluate {AutoDisym} and {ConsultMind} across psychiatry, respiratory medicine, fever clinics, and three public datasets. The results show that {AutoDisym} improves DSBN construction quality and that {ConsultMind} consistently improves diagnostic performance and explanation soundness. For example, \textsc{AutoDisym} achieves macro-averaged F1 scores of 81.37 for canonical symptoms and 72.19 for manifestations using GPT-5.6-Sol. {ConsultMind} improves Top-1 and Top-3 diagnostic accuracy by up to 22.15 and 37.89 percentage points, respectively. Physician evaluation further shows that {ConsultMind} improves ranking explanations, differential diagnoses, and diagnosis rationales across LLMs, resulting in more clinically sound diagnostic explanations. Our contributions are as follows:
\begin{itemize}[leftmargin=1.5em, itemsep=0.2em, topsep=0.2em, parsep=0pt, partopsep=0pt]
    \item We introduce \textsc{\textbf{AutoDisym}}, an automated pipeline that constructs Disorder--Symptom Bayesian Networks from diagnostic knowledge and clinical narratives, linking diagnostic hypotheses to potential inquiries.

    \item Building on these networks, we propose \textsc{\textbf{ConsultMind}}, an uncertainty-aware reasoning framework that translates evolving disorder posteriors into adaptive and interpretable consultation decisions.

    \item Across three clinical settings and external datasets, evaluations show that {AutoDisym} improves DSBN construction quality and that {ConsultMind} consistently improves diagnostic performance and explanation soundness.
\end{itemize}

\section{Related Work}

Medical knowledge has been incorporated into LLM-based diagnostic systems through several complementary paradigms~\cite{singhal2023large}. In-context approaches embed clinical guidelines~\cite{kresevic2024optimization,wang2024prompt} or diagnostic criteria directly into prompts~\cite{li2026care,savage2024diagnostic}, whereas retrieval-augmented generation retrieves case-relevant evidence from external medical corpora~\cite{xiong2024benchmarking,gaber2025evaluating}. Knowledge-graph-based retrieval further represents diseases~\cite{jia2025medikal}, symptoms~\cite{song2025graph}, examinations~\cite{gao2025large}, and treatments through structured relations~\cite{alber2025medical,zhou2026collaborative}, enabling relation-aware grounding~\cite{jiang2023graphcare}. However, existing systems exhibit rigid and insufficiently adaptive inquiry.

Automatic diagnostic consultation has evolved from structured control~\cite{sohn2026systematic} to learned and generative decision-making~\cite{xu2023medical}. Protocol-driven systems organize interactions~\cite{andreadis2024mixed} around clinical scales~\cite{sittig2024patient}, symptom checklists~\cite{ben2022assessing}, decision trees~\cite{you2023beyond}, and medical flowcharts~\cite{liu2026multi}. Reinforcement-learning and task-oriented dialogue methods~\cite{hou2023mtdiag} learn symptom-inquiry policies within predefined state and action spaces, whereas LLM-based systems generate follow-up questions and diagnoses directly from the dialogue context~\cite{tu2025towards}. However, LLM-based consultation decisions often remain opaque and difficult to interpret.

\section{\textsc{AutoDisym}}
\label{sec:autodisym}

\textsc{AutoDisym} uses specialized agents to construct a Disorder--Symptom Bayesian Network (DSBN) from diagnostic knowledge and labeled clinical narratives.

\paragraph{Data Preparation.}
\textsc{AutoDisym} takes two inputs: a curated diagnostic knowledge repository and a corpus of diagnosis-labeled clinical narratives. The repository contains diagnostic criteria from the International Classification of Diseases, 11th Revision (ICD-11), clinical guidelines, and selected consensus statements, with $\mathcal{K}_{\mathrm{ver}}$ denoting the guideline and consensus subset. The corpus includes consultation dialogues, electronic medical records, and case reports. Let $\mathcal{D}=\{d_k\}_{k=1}^{K}$ be the set of candidate disorders. We represent the corpus as $\mathcal{C}=\{(n_i,y_i)\}_{i=1}^{N}$, where $n_i$ is a clinical narrative and $y_i\in\mathcal{D}$ is its primary diagnosis. From these inputs, \textsc{AutoDisym} constructs the DSBN in four stages. Details are provided in Appendix.

\begin{figure}[t]
    \centering
    \includegraphics[width=1\columnwidth]{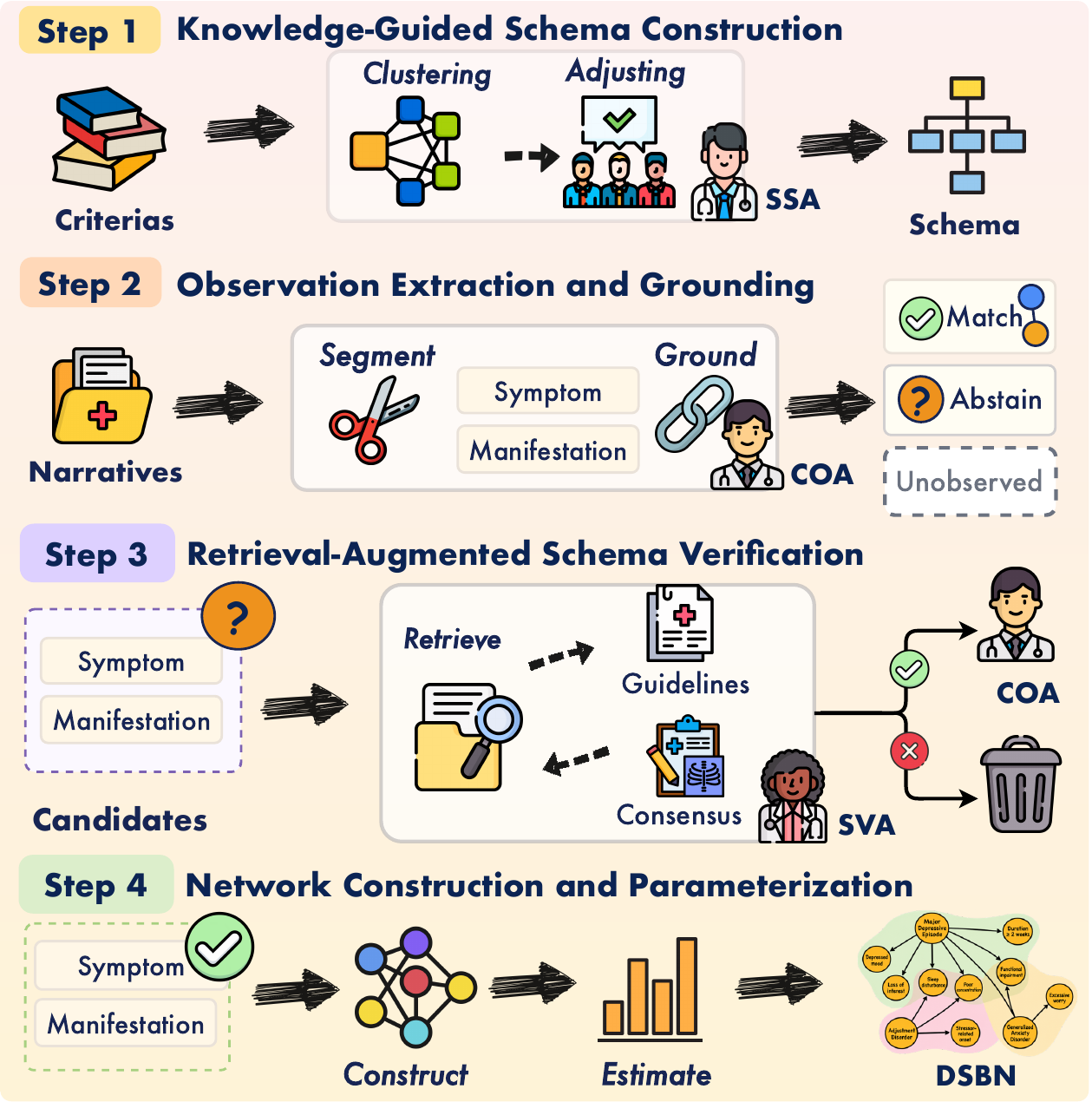}
    \caption{Overview of \textsc{AutoDisym}. An automated pipeline that constructs Disorder--Symptom Bayesian Networks (DSBN) from diagnostic knowledge and clinical narratives.}
    \label{fig:autodisym}
    \vspace{-8pt}
\end{figure}

\paragraph{Stage 1: Knowledge-Guided Schema Construction.}
For each disorder \(d_k\in\mathcal{D}\), the Symptom Schema Agent (SSA) processes its diagnostic criteria in two steps:

\begin{itemize}
    \item \textbf{Initialization:} SSA invokes \(\operatorname{Extract}(d_k)\) and dynamically orchestrates subagents to identify symptoms and their initial manifestations.

    \item \textbf{Schema refinement:} SSA clusters similar candidate expressions and assigns each cluster to an adjudication subagent, which \texttt{merges}, \texttt{retains}, or \texttt{removes}.
\end{itemize}

SSA consolidates these inventories into the canonical schema \(\mathcal{S}=\{S_j\}_{j=1}^{M}\), where each \(S_j\) has a manifestation state space \(\mathcal{V}_j\) (details see Appendix).

\paragraph{Stage 2: Observation Extraction and Grounding.}
The Clinical Observation Agent (COA) segments each narrative \(n_i\) and extracts symptom mentions, manifestations, and supporting spans. It retrieves related canonical symptoms and grounds each observation to a symptom--state pair, yielding three outcomes:

\begin{itemize}
    \item \textbf{Match:} Maps the observation to \((S_j,v)\), where \(v\in\mathcal{V}_j\).

    \item \textbf{Unobserved:} Leaves an unmentioned canonical symptom unassigned.

    \item \textbf{Abstention:} Defers an observation with no reliable mapping to the current schema.
\end{itemize}

COA retains each abstention and its supporting span as either a manifestation candidate \((S_j,\tilde{v})\) or a symptom candidate \((\tilde{S},\tilde{v})\) for schema verification.

\paragraph{Stage 3: Retrieval-Augmented Schema Verification.}
The Schema Verification Agent (SVA) verifies COA's manifestation and symptom candidates against \(\mathcal{K}_{\mathrm{ver}}\). For retrieval, \(\mathcal{K}_{\mathrm{ver}}\) is divided into fixed-size chunks \(\mathcal{B}_{\mathrm{ver}}\) and indexed by the BGE-M3~\citep{bge-m3} encoder \(f(\cdot)\). For each candidate \(c\), SVA builds a query \(q_c\) from the candidate and its supporting span, then retrieves the \(L\) most similar chunks:
\begin{equation}
    \mathcal{R}_c
    =
    \underset{p\in\mathcal{B}_{\mathrm{ver}}}
    {\operatorname{Top}\text{-}L}
    \operatorname{sim}\bigl(f(q_c),f(p)\bigr).
    \label{eq:rag-retrieval}
\end{equation}

\textit{Evidence-based schema expansion.}
Using \(\mathcal{R}_c\), SVA accepts candidates supported by clinical evidence and rejects the others. This process captures guideline- or consensus-supported symptoms and manifestations absent from the diagnostic criteria. SVA sends accepted candidates to SSA, which adds \(\tilde{v}\) to \(\mathcal{V}_j\) for \(c=(S_j,\tilde{v})\) or creates a symptom--state entry for \(c=(\tilde{S},\tilde{v})\). COA then re-grounds the affected observations and records each candidate's frequency. Details are provided in Appendix.

\paragraph{Stage 4: Network Construction and Parameterization.}
Finally, the canonical schema, grounded observations, and diagnosis labels define the DSBN. Let \(D\) denote the disorder variable and \(\mathbf{S}=(S_1,\ldots,S_M)\) the symptom variables, with edges \(D\rightarrow S_j\). The joint distribution factorizes as
\begin{equation}
    P(D,\mathbf{S})
    =
    P(D)\prod_{j=1}^{M}P(S_j\mid D).
    \label{eq:dsbn-factorization}
\end{equation}

The prior and conditional distributions are estimated from diagnosis labels and grounded observations using add-one smoothing (details see Appendix).

\section{\textsc{ConsultMind}}
\label{sec:consultmind}

\begin{figure*}
    \centering
    \includegraphics[width=1\textwidth]{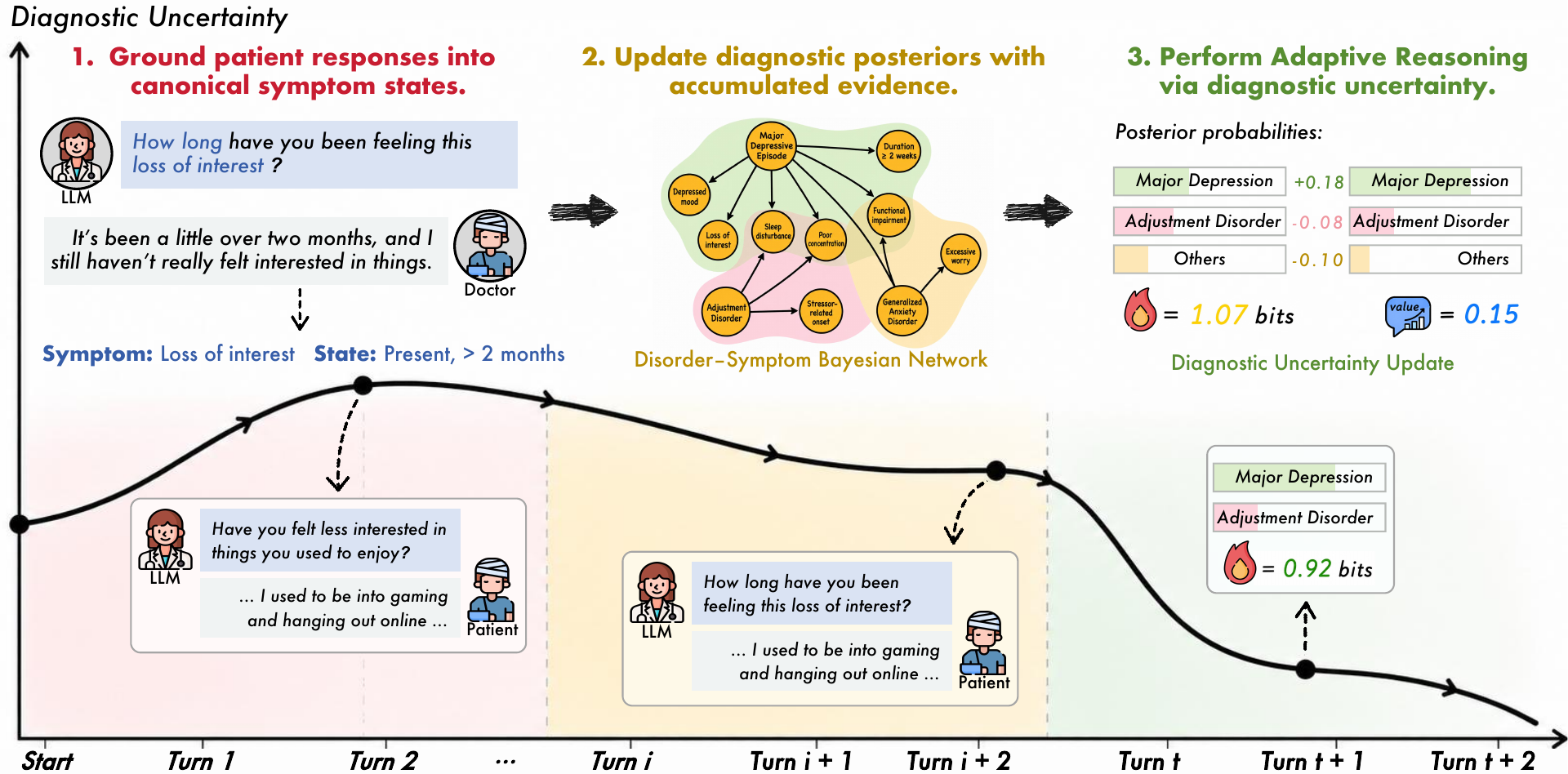}
    \caption{Automated Diagnostic Consultation via \textsc{ConsultMind}. It grounds patient responses into canonical symptom states, updates diagnostic posteriors through the DSBN, and uses posterior uncertainty to guide inquiries until reaching a diagnosis.}
    \label{fig:placeholder}
\end{figure*}

\textsc{ConsultMind} conducts sequential consultations using the DSBN constructed by AutoDisym. At each turn, \textit{Uncertainty Awareness} grounds the patient response and updates the diagnostic state, which \textit{Adaptive Reasoning} uses to continue the inquiry or make a diagnosis.

\subsection{Uncertainty Awareness}

Given a response \(r_t\) to a query about symptom \(S_{j_t}\), an LLM-based grounding module maps \(r_t\) to a canonical state \(v\in\mathcal{V}_{j_t}\). If no reliable mapping is supported, the module abstains without changing the evidence. Explicit negations map to \texttt{absent}, whereas unmentioned symptoms remain unobserved. Let \(\mathcal{O}_t\subseteq\{1,\ldots,M\}\) denote the symptoms observed by turn \(t\), with grounded evidence
\[
\mathbf{e}_t=\{S_j=e_{t,j}\mid j\in\mathcal{O}_t\}.
\]

For each binary disorder node \(D_k\), DSBN inference computes \(q_{t,k}=P(D_k=1\mid\mathbf{e}_t)\). Normalizing these values gives the diagnostic posterior:
\begin{equation}
    p_t(d_k)
    =
    \frac{q_{t,k}}
    {\sum_{\ell=1}^{K}q_{t,\ell}}.
    \label{eq:consultmind-posterior}
\end{equation}

Only grounded observations update the posterior; without evidence, it reduces to the normalized disorder priors. The remaining uncertainty is measured by posterior entropy:
\begin{equation}
    H_t
    =
    -\sum_{k=1}^{K}
    p_t(d_k)\log_2 p_t(d_k).
    \label{eq:posterior-entropy}
\end{equation}
here, \(H_t\) measures diagnostic uncertainty and informs subsequent reasoning.

\subsection{Adaptive Reasoning}

\textsc{ConsultMind} selects a strategy-specific inquiry target from the evolving diagnostic state. It encodes the  posterior and accumulated evidence into a six-dimensional state vector \(\mathbf{z}_t\), then compares it with prototypes for three strategies (details see Appendix):
\(\mathcal{A}=\{\mathrm{exp},\mathrm{diff},\mathrm{con}\}\), corresponding to \textit{Exploration}, \textit{Differentiation}, and \textit{Consolidation}. It computes weighted cosine similarities with additional emphasis on evidence density, normalizes them with softmax, and selects the highest-probability strategy \(a_t\in\mathcal{A}\). For symptom selection, let \(\mathcal{T}_t\) denote the top-\(N\) disorders, \(\mathcal{U}_t=\{1,\ldots,M\}\setminus\mathcal{O}_t\) the unobserved symptoms, and \(d_t^\star=\arg\max_{d\in\mathcal{D}}p_t(d)\) the leading disorder.

\paragraph{Exploration.}
Exploration prioritizes symptoms characteristic of at least one leading disorder relative to the others:
\begin{equation}
    s_{\mathrm{exp}}(j)
    =
    \max_{d\in\mathcal{T}_t}
    \frac{P(S_j\neq\mathrm{abs}\mid d)}
         {P(S_j\neq\mathrm{abs}\mid\neg d)}.
    \label{eq:exploration-score}
\end{equation}

\paragraph{Differentiation.}
Differentiation prioritizes symptoms with the greatest distributional differences among competing:
\begin{equation}
    s_{\mathrm{diff}}(j)
    =
    \operatorname*{mean}_{d\neq d'\in\mathcal{T}_t}
    \mathrm{JS}\!\left(
        \mathrm{CPD}_j(d),
        \mathrm{CPD}_j(d')
    \right).
    \label{eq:differentiation-score}
\end{equation}

\paragraph{Consolidation.}
Consolidation seeks further evidence for the leading disorder by prioritizing its likely symptoms:
\begin{equation}
    s_{\mathrm{con}}(j)
    =
    P(S_j\neq\mathrm{abs}\mid d_t^\star).
    \label{eq:consolidation-score}
\end{equation}
here, \(\mathrm{abs}\) is the absent state, and \(\mathrm{CPD}_j(d)\) is the categorical distribution of \(S_j\) under disorder \(d\). Given the selected strategy \(a_t\), the next inquiry target is
\begin{equation}
    j_{t+1}
    =
    \arg\max_{j\in\mathcal{U}_t}
    s_{a_t}(j).
    \label{eq:symptom-selection}
\end{equation}

An LLM converts \(S_{j_{t+1}}\) into a patient-facing question, whose response begins the next turn.

\paragraph{Termination.}
After each posterior update, ConsultMind decides whether to continue based on the current uncertainty and its reduction over a rolling window. The consultation ends when uncertainty is sufficiently low or no longer decreases; symptom exhaustion and the turn limit serve as fallback conditions (details see Appendix). It then returns the disorders ranked by posterior probability:
\begin{equation}
    \boldsymbol{\rho}_t
    =
    \operatorname{argsort}^{\downarrow}_{d\in\mathcal{D}}
    p_t(d).
    \label{eq:diagnostic-output}
\end{equation}

\section{Experiments}

\begin{table*}[t!]

\vspace{-0.5em}
\centering

\small
\setlength{\tabcolsep}{3pt}
\renewcommand{\arraystretch}{1.12}

\resizebox{\textwidth}{!}{
\begin{tabular}{l@{\hspace{12pt}}lll@{\hspace{12pt}}lll@{\hspace{12pt}}lll}
\toprule
& \multicolumn{3}{c}{\textbf{Psychiatry}}
& \multicolumn{3}{c}{\textbf{Respiratory Medicine}}
& \multicolumn{3}{c}{\textbf{Fever Clinic}} \\
\cmidrule(lr){2-4}
\cmidrule(lr){5-7}
\cmidrule(lr){8-10}
\textbf{Configuration}
& \textbf{Top-1} $\uparrow$
& \textbf{Top-3} $\uparrow$
& \textbf{MRR} $\uparrow$
& \textbf{Top-1} $\uparrow$
& \textbf{Top-3} $\uparrow$
& \textbf{MRR} $\uparrow$
& \textbf{Top-1} $\uparrow$
& \textbf{Top-3} $\uparrow$
& \textbf{MRR} $\uparrow$ \\
\midrule

\multicolumn{10}{l}{\rule[-0.7ex]{0pt}{3.0ex}\textbf{\textsc{Qwen3-8B}}} \\
\quad Direct
& 21.05\% & 35.79\% & 0.279
& 17.42\% & 31.64\% & 0.255
& 14.26\% & 26.91\% & 0.222 \\
\quad w/ RAG
& 28.31\%\ensuremath{_{\scriptsize +7.26}}
& 51.11\%\ensuremath{_{\scriptsize +15.32}}
& 0.383\ensuremath{_{\scriptsize +0.104}}
& 22.27\%\ensuremath{_{\scriptsize +4.85}}
& 42.56\%\ensuremath{_{\scriptsize +10.92}}
& 0.328\ensuremath{_{\scriptsize +0.073}}
& 19.01\%\ensuremath{_{\scriptsize +4.75}}
& 36.38\%\ensuremath{_{\scriptsize +9.47}}
& 0.284\ensuremath{_{\scriptsize +0.062}} \\
\quad w/ \textsc{ConsultMind}
& 41.80\%\ensuremath{_{\scriptsize +20.75}}
& 69.84\%\ensuremath{_{\scriptsize +34.05}}
& 0.538\ensuremath{_{\scriptsize +0.259}}
& 31.28\%\ensuremath{_{\scriptsize +13.86}}
& 55.91\%\ensuremath{_{\scriptsize +24.27}}
& 0.438\ensuremath{_{\scriptsize +0.183}}
& 27.84\%\ensuremath{_{\scriptsize +13.58}}
& 47.95\%\ensuremath{_{\scriptsize +21.04}}
& 0.376\ensuremath{_{\scriptsize +0.154}} \\

\cmidrule(lr){1-10}

\multicolumn{10}{l}{\rule[-0.7ex]{0pt}{3.0ex}\textbf{\textsc{Llama-3.1-8B}}} \\
\quad Direct
& 20.53\% & 40.53\% & 0.300
& 16.08\% & 35.87\% & 0.274
& 13.68\% & 30.42\% & 0.238 \\
\quad w/ RAG
& 28.28\%\ensuremath{_{\scriptsize +7.75}}
& 51.24\%\ensuremath{_{\scriptsize +10.71}}
& 0.388\ensuremath{_{\scriptsize +0.088}}
& 21.88\%\ensuremath{_{\scriptsize +5.80}}
& 43.77\%\ensuremath{_{\scriptsize +7.90}}
& 0.334\ensuremath{_{\scriptsize +0.060}}
& 19.03\%\ensuremath{_{\scriptsize +5.35}}
& 37.37\%\ensuremath{_{\scriptsize +6.95}}
& 0.289\ensuremath{_{\scriptsize +0.051}} \\
\quad w/ \textsc{ConsultMind}
& 42.68\%\ensuremath{_{\scriptsize +22.15}}
& 64.33\%\ensuremath{_{\scriptsize +23.80}}
& 0.521\ensuremath{_{\scriptsize +0.221}}
& 32.65\%\ensuremath{_{\scriptsize +16.57}}
& 53.42\%\ensuremath{_{\scriptsize +17.55}}
& 0.423\ensuremath{_{\scriptsize +0.149}}
& 28.96\%\ensuremath{_{\scriptsize +15.28}}
& 45.86\%\ensuremath{_{\scriptsize +15.44}}
& 0.365\ensuremath{_{\scriptsize +0.127}} \\

\cmidrule(lr){1-10}

\multicolumn{10}{l}{\rule[-0.7ex]{0pt}{3.0ex}\textbf{\textsc{Qwen3-32B}}} \\
\quad Direct
& 33.33\% & 40.32\% & 0.367
& 31.64\% & 36.91\% & 0.351
& 26.84\% & 33.28\% & 0.319 \\
\quad w/ RAG
& 39.75\%\ensuremath{_{\scriptsize +6.42}}
& 56.30\%\ensuremath{_{\scriptsize +15.98}}
& 0.470\ensuremath{_{\scriptsize +0.103}}
& 36.22\%\ensuremath{_{\scriptsize +4.58}}
& 48.91\%\ensuremath{_{\scriptsize +12.00}}
& 0.420\ensuremath{_{\scriptsize +0.069}}
& 31.07\%\ensuremath{_{\scriptsize +4.23}}
& 43.00\%\ensuremath{_{\scriptsize +9.72}}
& 0.376\ensuremath{_{\scriptsize +0.057}} \\
\quad w/ \textsc{ConsultMind}
& 51.67\%\ensuremath{_{\scriptsize +18.34}}
& 75.83\%\ensuremath{_{\scriptsize +35.51}}
& 0.625\ensuremath{_{\scriptsize +0.258}}
& 44.72\%\ensuremath{_{\scriptsize +13.08}}
& 63.58\%\ensuremath{_{\scriptsize +26.67}}
& 0.524\ensuremath{_{\scriptsize +0.173}}
& 38.92\%\ensuremath{_{\scriptsize +12.08}}
& 54.87\%\ensuremath{_{\scriptsize +21.59}}
& 0.461\ensuremath{_{\scriptsize +0.142}} \\

\cmidrule(lr){1-10}

\multicolumn{10}{l}{\rule[-0.7ex]{0pt}{3.0ex}\textbf{\textsc{Gemini-3.1-Pro}}} \\
\quad Direct
& 37.37\% & 40.00\% & 0.385
& 34.18\% & 37.26\% & 0.367
& 27.18\% & 33.54\% & 0.321 \\
\quad w/ RAG
& 42.34\%\ensuremath{_{\scriptsize +4.97}}
& 57.05\%\ensuremath{_{\scriptsize +17.05}}
& 0.483\ensuremath{_{\scriptsize +0.098}}
& 38.29\%\ensuremath{_{\scriptsize +4.11}}
& 49.48\%\ensuremath{_{\scriptsize +12.22}}
& 0.437\ensuremath{_{\scriptsize +0.070}}
& 31.19\%\ensuremath{_{\scriptsize +4.01}}
& 42.67\%\ensuremath{_{\scriptsize +9.13}}
& 0.374\ensuremath{_{\scriptsize +0.053}} \\
\quad w/ \textsc{ConsultMind}
& 51.58\%\ensuremath{_{\scriptsize +14.21}}
& 77.89\%\ensuremath{_{\scriptsize +37.89}}
& 0.629\ensuremath{_{\scriptsize +0.244}}
& 45.91\%\ensuremath{_{\scriptsize +11.73}}
& 64.42\%\ensuremath{_{\scriptsize +27.16}}
& 0.543\ensuremath{_{\scriptsize +0.176}}
& 38.64\%\ensuremath{_{\scriptsize +11.46}}
& 53.82\%\ensuremath{_{\scriptsize +20.28}}
& 0.454\ensuremath{_{\scriptsize +0.133}} \\

\cmidrule(lr){1-10}

\multicolumn{10}{l}{\rule[-0.7ex]{0pt}{3.0ex}\textbf{\textsc{DeepSeek-V4-Pro}}} \\
\quad Direct
& 35.48\% & 41.40\% & 0.382
& 32.03\% & 37.84\% & 0.359
& 27.92\% & 34.16\% & 0.329 \\
\quad w/ RAG
& 40.01\%\ensuremath{_{\scriptsize +4.53}}
& 56.40\%\ensuremath{_{\scriptsize +15.00}}
& 0.469\ensuremath{_{\scriptsize +0.087}}
& 35.65\%\ensuremath{_{\scriptsize +3.62}}
& 48.38\%\ensuremath{_{\scriptsize +10.54}}
& 0.416\ensuremath{_{\scriptsize +0.057}}
& 31.40\%\ensuremath{_{\scriptsize +3.48}}
& 42.61\%\ensuremath{_{\scriptsize +8.45}}
& 0.375\ensuremath{_{\scriptsize +0.046}} \\
\quad w/ \textsc{ConsultMind}
& 48.42\%\ensuremath{_{\scriptsize +12.94}}
& 74.74\%\ensuremath{_{\scriptsize +33.34}}
& 0.600\ensuremath{_{\scriptsize +0.218}}
& 42.36\%\ensuremath{_{\scriptsize +10.33}}
& 61.27\%\ensuremath{_{\scriptsize +23.43}}
& 0.501\ensuremath{_{\scriptsize +0.142}}
& 37.86\%\ensuremath{_{\scriptsize +9.94}}
& 52.94\%\ensuremath{_{\scriptsize +18.78}}
& 0.443\ensuremath{_{\scriptsize +0.114}} \\

\cmidrule(lr){1-10}

\multicolumn{10}{l}{\rule[-0.7ex]{0pt}{3.0ex}\textbf{\textsc{GPT-5.6-Sol}}} \\
\quad Direct
& 33.68\% & 38.42\% & 0.360
& 32.48\% & 38.76\% & 0.369
& 29.78\% & 36.42\% & 0.356 \\
\quad w/ RAG
& 39.58\%\ensuremath{_{\scriptsize +5.90}}
& 53.81\%\ensuremath{_{\scriptsize +15.39}}
& 0.456\ensuremath{_{\scriptsize +0.096}}
& 37.86\%\ensuremath{_{\scriptsize +5.38}}
& 50.65\%\ensuremath{_{\scriptsize +11.89}}
& 0.441\ensuremath{_{\scriptsize +0.072}}
& 34.85\%\ensuremath{_{\scriptsize +5.07}}
& 47.11\%\ensuremath{_{\scriptsize +10.69}}
& 0.424\ensuremath{_{\scriptsize +0.068}} \\
\quad w/ \textsc{ConsultMind}
& 50.53\%\ensuremath{_{\scriptsize +16.85}}
& 72.63\%\ensuremath{_{\scriptsize +34.21}}
& 0.599\ensuremath{_{\scriptsize +0.239}}
& 47.86\%\ensuremath{_{\scriptsize +15.38}}
& \textbf{65.18\%}\ensuremath{_{\scriptsize +26.42}}
& 0.549\ensuremath{_{\scriptsize +0.180}}
& \textbf{44.26\%}\ensuremath{_{\scriptsize +14.48}}
& \textbf{60.18\%}\ensuremath{_{\scriptsize +23.76}}
& \textbf{0.526}\ensuremath{_{\scriptsize +0.170}} \\

\cmidrule(lr){1-10}

\multicolumn{10}{l}{\rule[-0.7ex]{0pt}{3.0ex}\textbf{\textsc{Claude-Sonnet-5}}} \\
\quad Direct
& 34.74\% & 43.68\% & 0.390
& 33.27\% & 39.18\% & 0.374
& 29.13\% & 35.68\% & 0.344 \\
\quad w/ RAG
& 41.19\%\ensuremath{_{\scriptsize +6.45}}
& 60.02\%\ensuremath{_{\scriptsize +16.34}}
& 0.496\ensuremath{_{\scriptsize +0.106}}
& 38.54\%\ensuremath{_{\scriptsize +5.27}}
& 50.74\%\ensuremath{_{\scriptsize +11.56}}
& 0.445\ensuremath{_{\scriptsize +0.071}}
& 33.77\%\ensuremath{_{\scriptsize +4.64}}
& 45.47\%\ensuremath{_{\scriptsize +9.79}}
& 0.405\ensuremath{_{\scriptsize +0.061}} \\
\quad w/ \textsc{ConsultMind}
& \textbf{53.16\%}\ensuremath{_{\scriptsize +18.42}}
& \textbf{80.00\%}\ensuremath{_{\scriptsize +36.32}}
& \textbf{0.654}\ensuremath{_{\scriptsize +0.265}}
& \textbf{48.34\%}\ensuremath{_{\scriptsize +15.07}}
& 64.87\%\ensuremath{_{\scriptsize +25.69}}
& \textbf{0.552}\ensuremath{_{\scriptsize +0.178}}
& 42.38\%\ensuremath{_{\scriptsize +13.25}}
& 57.44\%\ensuremath{_{\scriptsize +21.76}}
& 0.497\ensuremath{_{\scriptsize +0.153}} \\

\bottomrule
\end{tabular}
}
\caption{Comparison of LLMs under direct diagnosis, diagnostic-criteria RAG, and \textsc{ConsultMind} across clinical specialties. Top-$k$ denotes diagnostic accuracy; MRR is the mean reciprocal rank of the correct diagnosis.}

\label{tab:consultmind_rag_estimated_results}
\vspace{-1em}
\end{table*}

\subsection{Data Collection}

We collected diagnosis-labeled cases from two tertiary hospitals and one center for disease control and prevention across different regions. The dataset comprises psychiatric consultation dialogues, respiratory electronic health records (EHRs), and clinical cases from fever clinics. Primary labels were derived from EHR discharge diagnoses.

\paragraph{Network Construction Corpus.}
We constructed a separate DSBN for each clinical setting from 14,541 cases: 4,102 respiratory, 4,644 psychiatric, and 5,795 fever-clinic cases. The psychiatric cases pair dialogue and record narratives, allowing cross-format consistency evaluation of AutoDisym diagnostic posteriors.

\paragraph{Evaluation Set.}
The disjoint evaluation set contains 990 cases spanning 78 disorders: 287 respiratory cases from 27 disorders, 312 psychiatric cases from 27 disorders, and 391 fever-clinic cases from 24 disorders.

\paragraph{Virtual Standard Patient Setup.}
We construct each clinical narrative-based virtual standardized patient (VSP) from a structured clinical record. At each turn, the VSP generates a patient response and cites its supporting record span. An independent auditor checks record consistency, evidence support, and unsupported clinical claims. For undocumented information, the VSP must express uncertainty rather than infer symptom absence. Responses that fail the audit are revised using its feedback until accepted. Details are provided in Appendix. Separate GPT-5.5 instances serve as the patient and auditor at temperature 0.

\paragraph{Ethics, Privacy, and Availability.}
Data processing followed an approved ethics protocol and the \textit{Declaration of Helsinki}. Evaluation records were institutionally de-identified and locally rewritten with Qwen3-32B to protect privacy. Clinicians verified that diagnostically relevant information was preserved. Further data and reproducibility details appear in the appendix. Institution names and ethics protocol identifiers are omitted for double-blind review. Upon acceptance, we will disclose them and release the evaluation set and learned DSBN parameters, while the source records used to construct the DSBN will remain restricted.

\paragraph{LLM Baselines.} We evaluate 10 LLMs in 3 groups: (1) Small LLMs: Qwen2.5-7B, Qwen3-8B and 32B~\citep{qwen3technicalreport}, Llama-3.1-8B~\citep{grattafiori2024llama}; (2) Large LLMs: GPT-5.6-Sol~\citep{openai2026gpt56}, Gemini-3.1-Pro~\citep{googledeepmind2026gemini31pro}, DeepSeek-V4-Pro~\citep{deepseek2026v4}, Claude-Sonnet-5~\citep{anthropic2026claudesonnet5}; and (3) Medical LLMs: ClinicalGPT-R1~\citep{lan2025clinicalgpt}, HuatuoGPT-O1-7B~\citep{chen2024huatuo}. Model configurations and inference details see  Appendix

\subsection{Experiment Results}

\begin{table}[t]
\centering

\resizebox{\columnwidth}{!}{%
\begin{tabular}{lccc}
\toprule
\textbf{Base Model} & \textbf{Cov. $\uparrow$} & \textbf{Prec. $\uparrow$} & \textbf{Halluc. $\downarrow$} \\
\midrule
QwQ-32B
    & 72.40 & 86.10 & 7.80 \\
\quad w/ Auditor
    & 88.91 {\scriptsize $(+16.51)$}
    & 93.27 {\scriptsize $(+7.17)$}
    & 3.08 {\scriptsize $(-4.72)$} \\

DeepSeek-V4-Flash
    & 76.85 & 88.74 & 6.42 \\
\quad w/ Auditor
    & 90.32 {\scriptsize $(+13.47)$}
    & 94.15 {\scriptsize $(+5.41)$}
    & 2.61 {\scriptsize $(-3.81)$} \\

GPT-5.5
    & 82.36 & 91.52 & 4.63 \\
\quad w/ Auditor
    & \textbf{92.18} {\scriptsize $(+9.82)$}
    & \textbf{95.06} {\scriptsize $(+3.54)$}
    & \textbf{1.94} {\scriptsize $(-2.69)$} \\
\bottomrule
\end{tabular}%
}

\caption{Projected VSP reliability with and without auditing.}
\label{tab:vsp-reliability}
\vspace{-8pt}

\end{table}

\paragraph{RQ1: Are Virtual Standardized Patients Reliable?}
We assess VSP reliability using three patient and auditor backbones on 60 cases with seven LLMs, yielding 1,260 dialogues. Fourteen crowdworkers double-annotated the sampled dialogues, with agreement exceeding $\kappa=0.85$. As shown in Table~\ref{tab:vsp-reliability}, audited VSPs achieve over 88.9\% coverage and 93.2\% precision, with hallucination below 3.1\% across all backbone sizes. Auditing improves coverage by up to 16.51 points and reduces hallucination by up to 4.72 points. These results demonstrate the reliability of VSPs across model capacities and the central role of the auditor.

\paragraph{RQ2: Does \textsc{ConsultMind} Improve Diagnostic Accuracy across LLMs and Specialties?}
We compare seven LLMs using direct prompting, RAG-enhanced prompting, and \textsc{ConsultMind} across psychiatry, respiratory medicine, and fever clinics. The RAG baseline follows a Naive RAG setup with 256-token chunks and Top-5 retrieval from the clinical materials used for DSBN construction, including diagnostic criteria, guidelines, and expert consensus (details see Appendix). As shown in Table~\ref{tab:consultmind_rag_estimated_results}, RAG improves over direct prompting, while \textsc{ConsultMind} further improves all 21 model--specialty combinations. Compared with direct prompting, the gains reach 22.15 percentage points in Top-1 accuracy, 37.89 points in Top-3 accuracy, and 0.265 in MRR. \texttt{Claude-Sonnet-5} performs best in psychiatry and is closely matched by \texttt{GPT-5.6-Sol} in respiratory medicine. \texttt{GPT-5.6-Sol} performs best in the fever clinic, reaching 44.26\% Top-1 accuracy, 60.18\% Top-3 accuracy, and 0.526 MRR. These results show that \textsc{ConsultMind} consistently outperforms both direct prompting and RAG-enhanced diagnosis across LLMs and specialties.

\begin{figure}[t]
    \centering
    \includegraphics[width=1\linewidth]{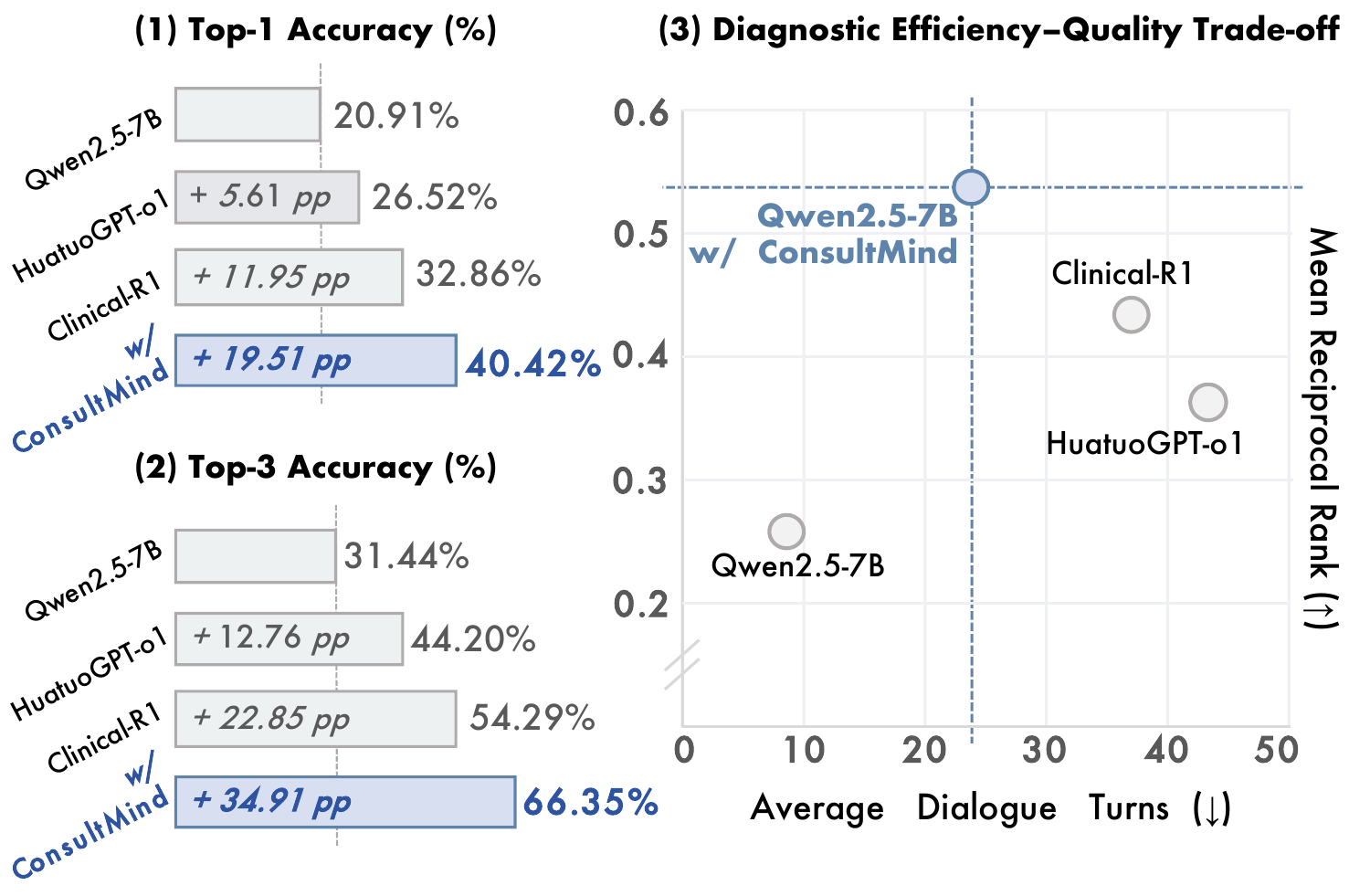}
    \caption{Comparison of \textsc{ConsultMind} with medical LLMs.}
    \label{fig:experiment_medical_llm_comparison}
    \vspace{-8pt}
\end{figure}

\paragraph{RQ3: How Does ConsultMind Compare with Medical LLMs?}
We compare ConsultMind with medically specialized LLMs built on the same Qwen2.5-7B base model. As shown in Figure~\ref{fig:experiment_medical_llm_comparison}, ConsultMind outperforms both medical LLMs across all diagnostic metrics. It achieves 40.42\% Top-1 and 66.35\% Top-3 accuracy, surpassing Clinical-R1 by 7.56 and 12.06 points, respectively. ConsultMind also achieves the highest MRR while requiring fewer dialogue turns than either medical LLM. These results show that ConsultMind improves the shared base model beyond specialized SFT.

\paragraph{RQ4: Does ConsultMind Generalize across Datasets?}
We evaluate ConsultMind on three external datasets: MentalHospital~\citep{yang2026mentalhospital}, MedSP1000~\citep{liang2026evaluating}, and AIHospital~\citep{fan2025ai}. The evaluation covers 310 psychiatric virtual patients from MentalHospital, 40 psychiatric and 48 respiratory cases from MedSP1000, and 14 fever-clinic cases from AIHospital. As shown in Table~\ref{tab:consultmind_cross_dataset}, ConsultMind improves all nine model and dataset combinations, with gains of up to 16.78 points in Top-1 accuracy, 24.51 points in Top-3 accuracy, and 0.187 in MRR. These consistent gains demonstrate robust generalization across external datasets and clinical settings.

\begin{table}[t]

\vspace{-0.5em}
\centering

\small
\setlength{\tabcolsep}{4pt}
\renewcommand{\arraystretch}{0.9}

\resizebox{\columnwidth}{!}{
\begin{tabular}{l@{\hspace{16pt}}lll}
\toprule
\textbf{Model}
& \textbf{Top-1} $\uparrow$
& \textbf{Top-3} $\uparrow$
& \textbf{MRR} $\uparrow$ \\
\midrule

\multicolumn{4}{@{}l@{}}{
\makebox[\columnwidth]{
\rule[-0.7ex]{0pt}{3.0ex}
\textsc{\textbf{MentalHospital}}
\hfill
Psychiatry
}} \\
\cmidrule(lr){1-4}

Llama-3.1-8B
& 16.77\% & 34.19\% & 0.249 \\
\quad w/ \textsc{ConsultMind}
& 33.55\%\ensuremath{_{\scriptsize +16.78}}
& 51.61\%\ensuremath{_{\scriptsize +17.42}}
& 0.430\ensuremath{_{\scriptsize +0.181}} \\
\cmidrule(lr){1-4}

Qwen3-32B
& 27.74\% & 35.81\% & 0.325 \\
\quad w/ \textsc{ConsultMind}
& 40.65\%\ensuremath{_{\scriptsize +12.91}}
& 60.32\%\ensuremath{_{\scriptsize +24.51}}
& 0.512\ensuremath{_{\scriptsize +0.187}} \\
\cmidrule(lr){1-4}

GPT-5.6-Sol
& 29.03\% & 34.19\% & 0.329 \\
\quad w/ \textsc{ConsultMind}
& 39.68\%\ensuremath{_{\scriptsize +10.65}}
& 56.13\%\ensuremath{_{\scriptsize +21.94}}
& 0.487\ensuremath{_{\scriptsize +0.158}} \\

\midrule

\multicolumn{4}{@{}l@{}}{
\makebox[\columnwidth]{
\rule[-0.7ex]{0pt}{3.0ex}
\textsc{\textbf{MedSP1000}}
\hfill
Psychiatry, Respiratory medicine
}} \\
\cmidrule(lr){1-4}

Llama-3.1-8B
& 14.77\% & 32.95\% & 0.230 \\
\quad w/ \textsc{ConsultMind}
& 27.27\%\ensuremath{_{\scriptsize +12.50}}
& 47.73\%\ensuremath{_{\scriptsize +14.78}}
& 0.375\ensuremath{_{\scriptsize +0.145}} \\
\cmidrule(lr){1-4}

Qwen3-32B
& 25.00\% & 34.09\% & 0.301 \\
\quad w/ \textsc{ConsultMind}
& 35.23\%\ensuremath{_{\scriptsize +10.23}}
& 53.41\%\ensuremath{_{\scriptsize +19.32}}
& 0.450\ensuremath{_{\scriptsize +0.149}} \\
\cmidrule(lr){1-4}

GPT-5.6-Sol
& 27.27\% & 32.95\% & 0.312 \\
\quad w/ \textsc{ConsultMind}
& 37.50\%\ensuremath{_{\scriptsize +10.23}}
& 51.14\%\ensuremath{_{\scriptsize +18.19}}
& 0.452\ensuremath{_{\scriptsize +0.140}} \\

\midrule

\multicolumn{4}{@{}l@{}}{
\makebox[\columnwidth]{
\rule[-0.7ex]{0pt}{3.0ex}
\textsc{\textbf{AIHospital}}
\hfill
Fever Clinic
}} \\
\cmidrule(lr){1-4}

Llama-3.1-8B
& 21.43\% & 35.71\% & 0.280 \\
\quad w/ \textsc{ConsultMind}
& 35.71\%\ensuremath{_{\scriptsize +14.28}}
& 50.00\%\ensuremath{_{\scriptsize +14.29}}
& 0.433\ensuremath{_{\scriptsize +0.153}} \\
\cmidrule(lr){1-4}

Qwen3-32B
& 28.57\% & 35.71\% & 0.327 \\
\quad w/ \textsc{ConsultMind}
& 42.86\%\ensuremath{_{\scriptsize +14.29}}
& 57.14\%\ensuremath{_{\scriptsize +21.43}}
& 0.510\ensuremath{_{\scriptsize +0.183}} \\
\cmidrule(lr){1-4}

GPT-5.6-Sol
& 28.57\% & 35.71\% & 0.330 \\
\quad w/ \textsc{ConsultMind}
& 42.86\%\ensuremath{_{\scriptsize +14.29}}
& 57.14\%\ensuremath{_{\scriptsize +21.43}}
& 0.507\ensuremath{_{\scriptsize +0.177}} \\

\bottomrule
\end{tabular}
}
\caption{Diagnostic performance of LLMs with and without \textsc{ConsultMind} across datasets.}
\label{tab:consultmind_cross_dataset}
\end{table}

\paragraph{RQ5: Does ConsultMind Improve the Clinical Soundness of Diagnostic Explanations?}

\begin{figure}[h]
    \centering
    \includegraphics[width=0.9\linewidth]{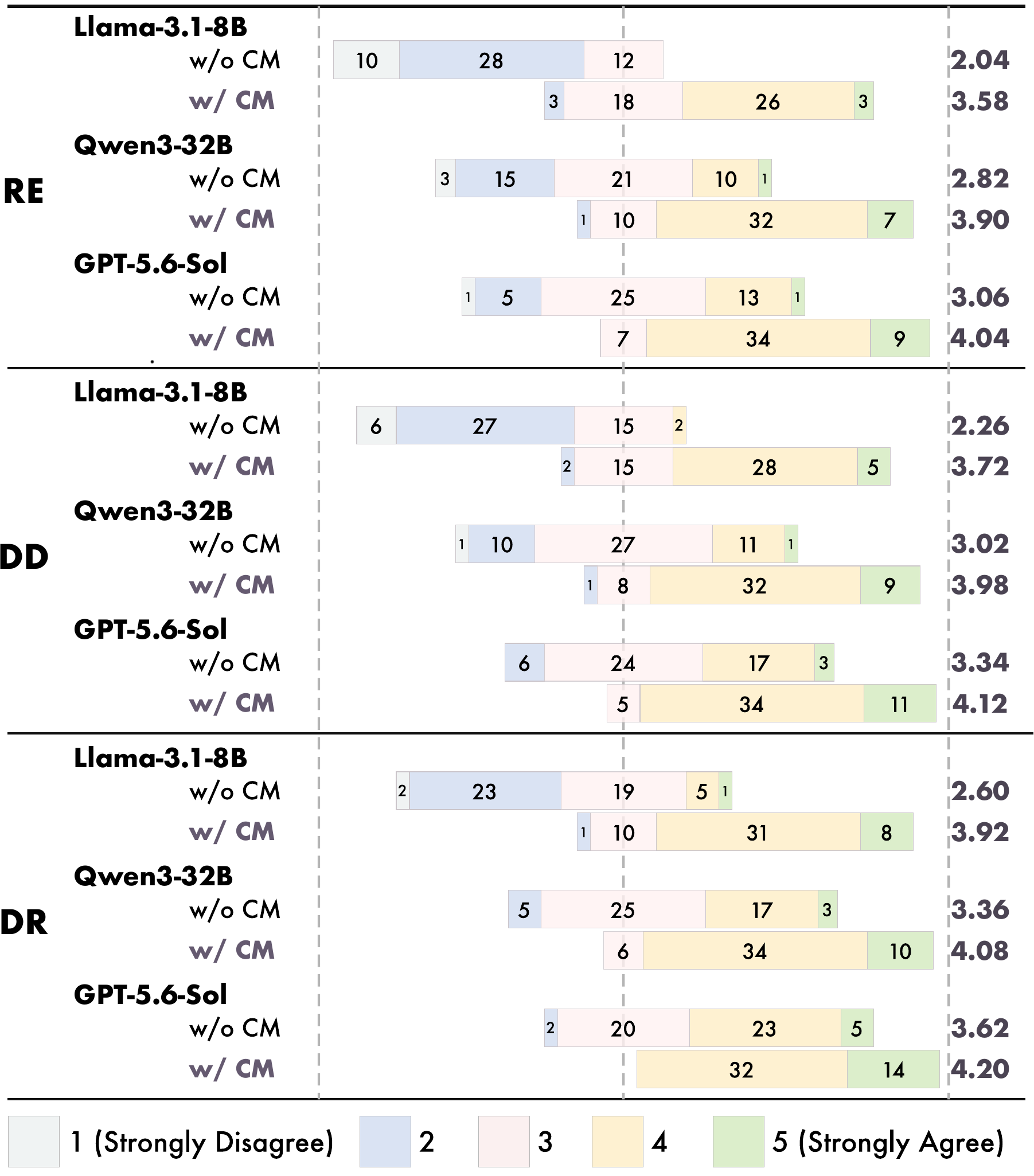}
    \caption{Physician ratings of diagnostic explanation quality.}
    \label{fig:explanation_quality}
\end{figure}

\begin{figure}[t!]
    \centering
    \includegraphics[width=1.0\linewidth]{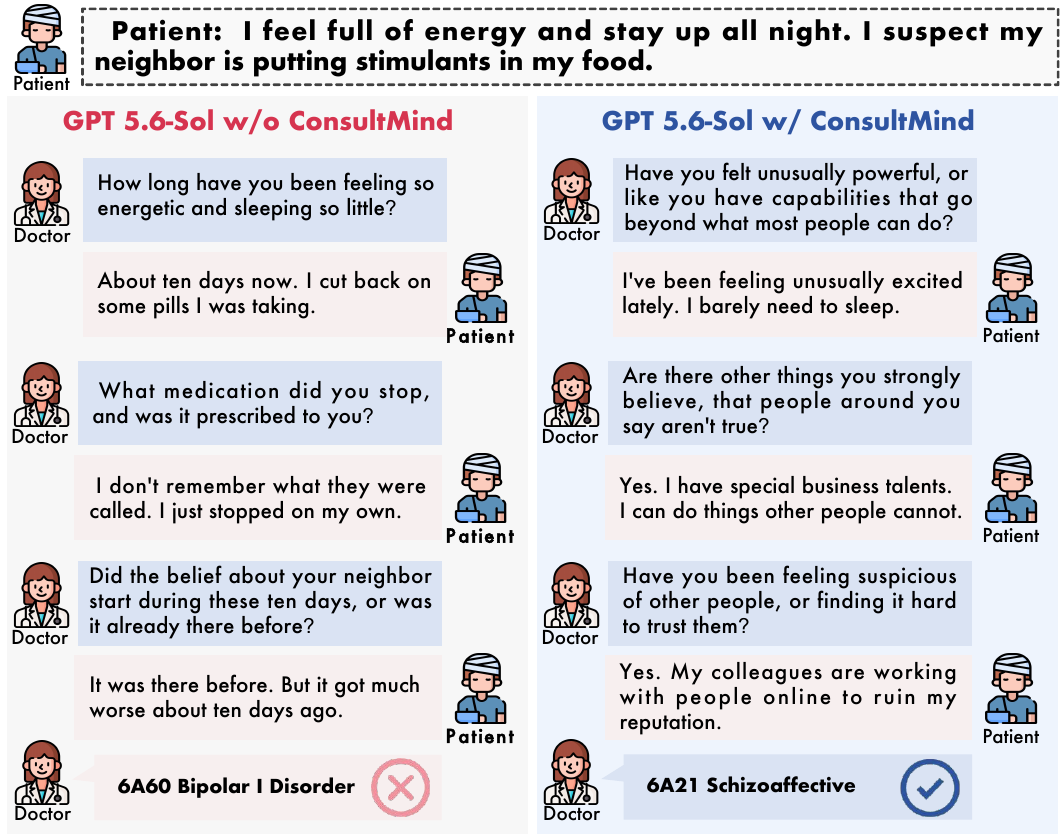}
    \caption{Consult comparison with/without \textsc{ConsultMind}.}
    \label{fig:case_study}
    \vspace{-12pt}
\end{figure}

We compare paired explanations from three representative LLMs on 50 correctly diagnosed cases per model, using their base and ConsultMind-enhanced versions. Each explanation follows a three-part template: Ranking Explanation (RE), Differential Diagnosis (DD), and Diagnosis Rationale (DR). Specialists from the corresponding departments rate the anonymized and shuffled explanations on a five-point scale (details see Appendix). As shown in Figure~\ref{fig:explanation_quality}, ConsultMind improves models, with the largest overall gain for Llama-3.1-8B from 2.30 to 3.74. Its RE score shows the largest component gain, increasing from 2.04 to 3.58. These results demonstrate that ConsultMind produces more clinically sound diagnostic explanations.

\paragraph{RQ6: How Reliable Are AutoDisym-Generated Symptom Schemas?}

We compare AutoDisym-generated schemas with expert gold schemas for six disorders across three departments: \textit{Influenza}, \textit{Dengue fever}, \textit{Emphysema}, \textit{Acute bronchiolitis}, \textit{Bipolar type II disorder}, and \textit{Recurrent depressive disorder}. Two experts independently annotate canonical symptoms, normalized manifestations, and their pairs from the same diagnostic materials, while a third resolves disagreements (details see Appendix). Using the same backbones, we compare Direct LLM, RAG-enhanced LLM, and AutoDisym. Direct LLM uses parametric knowledge, while the other methods share a Naive RAG setting with 256-token chunks and Top-5 retrieval. After synonym normalization, one-to-one matches are correct only when both the symptom and manifestation match the gold schema above 0.85 similarity. Table~\ref{tab:autodisym_schema_agreement} reports disorder-level macro precision, recall, and F1. AutoDisym performs best with both backbones, reaching 81.37 symptom F1 and 72.19 manifestation F1 with GPT-5.6-Sol. It also reduces the manifestation-F1 gap between backbones from 9.68 to 5.96, demonstrating more reliable schemas and less dependence on backbone capability.

\begin{table}[t]

\centering

\small
\setlength{\tabcolsep}{3pt}
\renewcommand{\arraystretch}{1.0}

\resizebox{\columnwidth}{!}{
\begin{tabular}{lcccccc}
\toprule
& \multicolumn{3}{c}{\textbf{Symptoms}}
& \multicolumn{3}{c}{\textbf{Manifestations}} \\
\cmidrule(lr){2-4}
\cmidrule(lr){5-7}
\textbf{Configuration}
& \textbf{Precision} $\uparrow$
& \textbf{Recall} $\uparrow$
& \textbf{F1} $\uparrow$
& \textbf{Precision} $\uparrow$
& \textbf{Recall} $\uparrow$
& \textbf{F1} $\uparrow$ \\
\midrule

\multicolumn{7}{l}{\rule[-0.7ex]{0pt}{3.0ex}\textbf{Qwen3-32B}} \\
\cmidrule(lr){1-7}

\quad Direct
& 52.17 & 42.46 & 46.82
& 40.86 & 29.36 & 34.17 \\

\quad w/ RAG
& 67.69 & 61.34 & 64.36
& 57.89 & 49.54 & 53.39 \\

\quad \textbf{w/ \textsc{AutoDisym}}
& \textbf{76.78} & \textbf{72.75} & \textbf{74.71}
& \textbf{69.46} & \textbf{63.28} & \textbf{66.23} \\

\midrule

\multicolumn{7}{l}{\rule[-0.7ex]{0pt}{3.0ex}\textbf{GPT-5.6-Sol}} \\
\cmidrule(lr){1-7}

\quad Direct
& 60.66 & 54.78 & 57.57
& 48.62 & 39.94 & 43.85 \\

\quad w/ RAG
& 74.14 & 69.04 & 71.50
& 65.22 & 55.54 & 59.99 \\

\quad \textbf{w/ \textsc{AutoDisym}}
& \textbf{84.34} & \textbf{78.61} & \textbf{81.37}
& \textbf{75.91} & \textbf{68.81} & \textbf{72.19} \\

\bottomrule
\end{tabular}
}
\caption{Schema agreement across configurations.}
\label{tab:autodisym_schema_agreement}
\vspace{-8pt}
\end{table}

\vspace{-4pt}
 
\subsection{Case Study}

Figure~\ref{fig:case_study} compares GPT-5.6-Sol with and without ConsultMind on a patient reporting increased energy, reduced sleep, and persecutory beliefs. Without ConsultMind, the model follows the medication-related narrative and predicts bipolar I disorder. With ConsultMind, it probes grandiosity, fixed false beliefs, and suspiciousness, yielding stronger evidence for schizoaffective disorder. This case shows that ConsultMind reduces narrative drift by directing inquiry toward discriminative symptoms in overlapping presentations.

\section{Conclusion}

We introduced \textsc{AutoDisym}, an automated pipeline for constructing Disorder--Symptom Bayesian Networks (DSBNs), and \textsc{ConsultMind}, an uncertainty-aware framework for sequential diagnostic consultation. {AutoDisym} integrates diagnostic knowledge with diagnosis-labeled clinical narratives, while {ConsultMind} updates disorder posteriors and uses uncertainty to guide inquiry and diagnosis. Evaluations across three clinical settings and three public datasets show that {AutoDisym} constructs high-quality DSBNs and {ConsultMind} consistently improves diagnostic performance and explanation soundness. {ConsultMind} increases Top-1 and Top-3 accuracy by up to 22.15 and 37.89 percentage points, respectively. Physician evaluation further confirms improvements in ranking explanation, differential diagnosis, and diagnosis rationale across LLMs of different scales. This work offers a promising approach to automatic diagnostic consultation.


\bibliography{aaai2027}

@article{kresevic2024optimization,
  title={Optimization of hepatological clinical guidelines interpretation by large language models: a retrieval augmented generation-based framework},
  author={Kresevic, Simone and Giuffr{\`e}, Mauro and Ajcevic, Milos and Accardo, Agostino and Croc{\`e}, Lory S and Shung, Dennis L},
  journal={NPJ digital medicine},
  volume={7},
  number={1},
  pages={102},
  year={2024},
  publisher={Nature Publishing Group UK London}
}

@article{wang2024prompt,
  title={Prompt engineering in consistency and reliability with the evidence-based guideline for LLMs},
  author={Wang, Li and Chen, Xi and Deng, XiangWen and Wen, Hao and You, MingKe and Liu, WeiZhi and Li, Qi and Li, Jian},
  journal={NPJ digital medicine},
  volume={7},
  number={1},
  pages={41},
  year={2024},
  publisher={Nature Publishing Group UK London}
}

@article{li2026care,
  title={CARE: A Clinical Agentic Reasoning Engine to Enhance Real-World Diagnostic Accuracy via Structured Medical Reasoning},
  author={Li, Wenjie and Zhang, Yujie and Wang, Chenrun and Li, Yueqi and He, Xingqi and Wang, Angela Lin and Xu, Mengzhe and Zhang, Fanrui and Sun, Haoran and Wang, Kailing and others},
  journal={Expert Systems with Applications},
  pages={131476},
  year={2026},
  publisher={Elsevier}
}

@article{savage2024diagnostic,
  title={Diagnostic reasoning prompts reveal the potential for large language model interpretability in medicine},
  author={Savage, Thomas and Nayak, Ashwin and Gallo, Robert and Rangan, Ekanath and Chen, Jonathan H},
  journal={NPJ Digital Medicine},
  volume={7},
  number={1},
  pages={20},
  year={2024},
  publisher={Nature Publishing Group UK London}
}

@inproceedings{xiong2024benchmarking,
  title={Benchmarking retrieval-augmented generation for medicine},
  author={Xiong, Guangzhi and Jin, Qiao and Lu, Zhiyong and Zhang, Aidong},
  booktitle={Findings of the Association for Computational Linguistics: ACL 2024},
  pages={6233--6251},
  year={2024}
}

@article{gaber2025evaluating,
  title={Evaluating large language model workflows in clinical decision support for triage and referral and diagnosis},
  author={Gaber, Farieda and Shaik, Maqsood and Allega, Fabio and Bilecz, Agnes Julia and Busch, Felix and Goon, Kelsey and Franke, Vedran and Akalin, Altuna},
  journal={npj Digital Medicine},
  volume={8},
  number={1},
  pages={263},
  year={2025},
  publisher={Nature Publishing Group UK London}
}

@inproceedings{jia2025medikal,
  title={medikal: Integrating knowledge graphs as assistants of llms for enhanced clinical diagnosis on emrs},
  author={Jia, Mingyi and Duan, Junwen and Song, Yan and Wang, Jianxin},
  booktitle={Proceedings of the 31st International Conference on Computational Linguistics},
  pages={9278--9298},
  year={2025}
}

@article{song2025graph,
  title={Graph retrieval augmented large language models for facial phenotype associated rare genetic disease},
  author={Song, Jie and Xu, Zhichuan and He, Mengqiao and Feng, Jinhua and Shen, Bairong},
  journal={NPJ digital medicine},
  volume={8},
  number={1},
  pages={543},
  year={2025},
  publisher={Nature Publishing Group UK London}
}

@article{gao2025large,
  title={Large language model powered knowledge graph construction for mental health exploration},
  author={Gao, Shan and Yu, Kaixian and Yang, Yue and Yu, Sheng and Shi, Chenglong and Wang, Xueqin and Tang, Niansheng and Zhu, Hongtu},
  journal={Nature Communications},
  volume={16},
  number={1},
  pages={7526},
  year={2025},
  publisher={Nature Publishing Group UK London}
}

@article{alber2025medical,
  title={Medical large language models are vulnerable to data-poisoning attacks},
  author={Alber, Daniel Alexander and Yang, Zihao and Alyakin, Anton and Yang, Eunice and Rai, Sumedha and Valliani, Aly A and Zhang, Jeff and Rosenbaum, Gabriel R and Amend-Thomas, Ashley K and Kurland, David B and others},
  journal={Nature Medicine},
  volume={31},
  number={2},
  pages={618--626},
  year={2025},
  publisher={Nature Publishing Group US New York}
}

@article{zhou2026collaborative,
  title={A collaborative large language model for drug analysis},
  author={Zhou, Hongjian and Liu, Fenglin and Wu, Jinge and Zhang, Wenjun and Huang, Guowei and Clifton, Lei and Eyre, David and Luo, Haochen and Liu, Fengyuan and Branson, Kim and others},
  journal={Nature Biomedical Engineering},
  volume={10},
  number={5},
  pages={870--881},
  year={2026},
  publisher={Nature Publishing Group UK London}
}

@inproceedings{jiang2023graphcare,
  title={Graphcare: Enhancing healthcare predictions with personalized knowledge graphs},
  author={Jiang, Pengcheng and Xiao, Cao and Cross, Adam Richard and Sun, Jimeng},
  booktitle={The Twelfth International Conference on Learning Representations},
  year={2023}
}

@article{sohn2026systematic,
  title={Systematic review and meta analysis of chatbots in the management of depressive and anxiety symptoms},
  author={Sohn, Jun-Seok and Ha, Byeong-Gwan and Park, SoHyun and Kim, Jae-Jin and Lee, Eojin and Oh, Hyangkyeong and Lee, San and Kim, Eunjoo},
  journal={NPJ Digital Medicine},
  year={2026},
  publisher={Nature Publishing Group UK London}
}

@inproceedings{xu2023medical,
  title={Medical dialogue generation via dual flow modeling},
  author={Xu, Kaishuai and Hou, Wenjun and Cheng, Yi and Wang, Jian and Li, Wenjie},
  booktitle={Findings of the Association for Computational Linguistics: ACL 2023},
  pages={6771--6784},
  year={2023}
}

@article{andreadis2024mixed,
  title={Mixed methods assessment of the influence of demographics on medical advice of ChatGPT},
  author={Andreadis, Katerina and Newman, Devon R and Twan, Chelsea and Shunk, Amelia and Mann, Devin M and Stevens, Elizabeth R},
  journal={Journal of the American Medical Informatics Association},
  volume={31},
  number={9},
  pages={2002--2009},
  year={2024},
  publisher={Oxford University Press}
}

@article{sittig2024patient,
  title={Patient-centered clinical decision support challenges and opportunities identified from workflow execution models},
  author={Sittig, Dean F and Boxwala, Aziz and Wright, Adam and Zott, Courtney and Gauthreaux, Nicole A and Swiger, James and Lomotan, Edwin A and Dullabh, Prashila},
  journal={Journal of the American Medical Informatics Association},
  volume={31},
  number={8},
  pages={1682--1692},
  year={2024},
  publisher={Oxford University Press}
}

@article{ben2022assessing,
  title={Assessing data gathering of chatbot based symptom checkers-a clinical vignettes study},
  author={Ben-Shabat, Niv and Sharvit, Gal and Meimis, Ben and Joya, Daniel Ben and Sloma, Ariel and Kiderman, David and Shabat, Aviv and Tsur, Avishai M and Watad, Abdulla and Amital, Howard},
  journal={International Journal of Medical Informatics},
  volume={168},
  pages={104897},
  year={2022},
  publisher={Elsevier}
}

@article{you2023beyond,
  title={Beyond self-diagnosis: how a chatbot-based symptom checker should respond},
  author={You, Yue and Tsai, Chun-Hua and Li, Yao and Ma, Fenglong and Heron, Christopher and Gui, Xinning},
  journal={ACM Transactions on Computer-Human Interaction},
  volume={30},
  number={4},
  pages={1--44},
  year={2023},
  publisher={ACM New York, NY, USA}
}

@article{liu2026multi,
  title={A multi-agent framework combining large language models with medical flowcharts for self-triage},
  author={Liu, Yujia and Yu, Sophia and Jin, Hongyue and Wen, Jessica and Qian, Alexander and Lee, Terrence and Ramsis, Mattheus and Choi, Gi Won and Qin, Lianhui and Liu, Xin and others},
  journal={Nature Health},
  pages={1--10},
  year={2026},
  publisher={Nature Publishing Group UK London}
}

@inproceedings{hou2023mtdiag,
  title={Mtdiag: an effective multi-task framework for automatic diagnosis},
  author={Hou, Zhenyu and Cen, Yukuo and Liu, Ziding and Wu, Dongxue and Wang, Baoyan and Li, Xuanhe and Hong, Lei and Tang, Jie},
  booktitle={Proceedings of the AAAI Conference on Artificial Intelligence},
  volume={37},
  number={12},
  pages={14241--14248},
  year={2023}
}

@article{tu2025towards,
  title={Towards conversational diagnostic artificial intelligence},
  author={Tu, Tao and Schaekermann, Mike and Palepu, Anil and Saab, Khaled and Freyberg, Jan and Tanno, Ryutaro and Wang, Amy and Li, Brenna and Amin, Mohamed and Cheng, Yong and others},
  journal={Nature},
  volume={642},
  number={8067},
  pages={442--450},
  year={2025},
  publisher={Nature Publishing Group UK London}
}

@article{li2024mediq,
  title={Mediq: Question-asking llms and a benchmark for reliable interactive clinical reasoning},
  author={Li, Shuyue S and Balachandran, Vidhisha and Feng, Shangbin and Ilgen, Jonathan S and Pierson, Emma and Koh, Pang W and Tsvetkov, Yulia},
  journal={Advances in Neural Information Processing Systems},
  volume={37},
  pages={28858--28888},
  year={2024}
}

@article{johri2025evaluation,
  title={An evaluation framework for clinical use of large language models in patient interaction tasks},
  author={Johri, Shreya and Jeong, Jaehwan and Tran, Benjamin A and Schlessinger, Daniel I and Wongvibulsin, Shannon and Barnes, Leandra A and Zhou, Hong-Yu and Cai, Zhuo Ran and Van Allen, Eliezer M and Kim, David and others},
  journal={Nature medicine},
  volume={31},
  number={1},
  pages={77--86},
  year={2025},
  publisher={Nature Publishing Group US New York}
}

@article{werthaim2026benchmark,
  title={A benchmark for evaluating diagnostic questioning efficiency of LLMs in patient conversations},
  author={Werthaim, Mai and Kimhi, Maya and Apartsin, Alexander and Aperstein, Yehudit},
  journal={Scientific Reports},
  year={2026},
  publisher={Nature Publishing Group UK London}
}

@misc{gong2025dialogue,
      title={The Dialogue That Heals: A Comprehensive Evaluation of Doctor Agents' Inquiry Capability}, 
      author={Linlu Gong and Ante Wang and Yunghwei Lai and Weizhi Ma and Yang Liu},
      year={2025},
      eprint={2509.24958},
      archivePrefix={arXiv},
      primaryClass={cs.CL},
      url={https://arxiv.org/abs/2509.24958}, 
}

@misc{sanghvi2026medxagent,
      title={MeDxAgent: Multi-Agent Consultation for Interactive Medical Diagnosis}, 
      author={Akshat Sanghvi and Naren Akash and Raza Imam and Amit Sharma and Mohit Jain},
      year={2026},
      eprint={2606.03416},
      archivePrefix={arXiv},
      primaryClass={cs.MA},
      url={https://arxiv.org/abs/2606.03416}, 
}

@article{yuan2024efficient,
  title={Efficient symptom inquiring and diagnosis via adaptive alignment of reinforcement learning and classification},
  author={Yuan, Hongyi and Yu, Sheng},
  journal={Artificial Intelligence in Medicine},
  volume={148},
  pages={102748},
  year={2024},
  publisher={Elsevier}
}

@inproceedings{sun2026mentalseek,
  title={MentalSeek-Dx: Towards Progressive Hypothetico-Deductive Reasoning for Real-world Psychiatric Diagnosis},
  author={Sun, Xiao and Yang, Yuming and Jiang, Xinyi and Tian, Yu and Zhu, Junnan and Zhong, Jiang and Lei, Qin and Huang, Jingwang and Zeng, Haoyang and Zhou, Xinyu and others},
  booktitle={Proceedings of the 64th Annual Meeting of the Association for Computational Linguistics (Volume 1: Long Papers)},
  pages={26600--26636},
  year={2026}
}

@article{saab2026advancing,
  title={Advancing conversational diagnostic AI with multimodal reasoning},
  author={Saab, Khaled and Park, Chunjong and Strother, Tim and Freyberg, Jan and Barrett, David GT and Cheng, Yong and Weng, Wei-Hung and Stutz, David and Tomasev, Nenad and Palepu, Anil and others},
  journal={Nature Medicine},
  pages={1--11},
  year={2026},
  publisher={Nature Publishing Group US New York}
}

@misc{tu2024towards,
      title={Towards Conversational Diagnostic AI}, 
      author={Tao Tu and Anil Palepu and Mike Schaekermann and Khaled Saab and Jan Freyberg and Ryutaro Tanno and Amy Wang and Brenna Li and Mohamed Amin and Nenad Tomasev and Shekoofeh Azizi and Karan Singhal and Yong Cheng and Le Hou and Albert Webson and Kavita Kulkarni and S Sara Mahdavi and Christopher Semturs and Juraj Gottweis and Joelle Barral and Katherine Chou and Greg S Corrado and Yossi Matias and Alan Karthikesalingam and Vivek Natarajan},
      year={2024},
      eprint={2401.05654},
      archivePrefix={arXiv},
      primaryClass={cs.AI},
      url={https://arxiv.org/abs/2401.05654}, 
}

@inproceedings{ren2025diallms,
  title={DiaLLMs: EHR-Enhanced Clinical Conversational System for Clinical Test Recommendation and Diagnosis Prediction},
  author={Ren, Weijieying and Zhao, Tianxiang and Wang, Lei and Wang, Tianchun and Honavar, Vasant G},
  booktitle={Findings of the Association for Computational Linguistics: ACL 2025},
  pages={25622--25635},
  year={2025}
}

@article{singhal2023large,
  title={Large language models encode clinical knowledge},
  author={Singhal, Karan and Azizi, Shekoofeh and Tu, Tao and Mahdavi, S Sara and Wei, Jason and Chung, Hyung Won and Scales, Nathan and Tanwani, Ajay and Cole-Lewis, Heather and Pfohl, Stephen and others},
  journal={Nature},
  volume={620},
  number={7972},
  pages={172--180},
  year={2023},
  publisher={Nature Publishing Group UK London}
}

@inproceedings{zeng2020meddialog,
  title={MedDialog: Large-scale medical dialogue datasets},
  author={Zeng, Guangtao and Yang, Wenmian and Ju, Zeqian and Yang, Yue and Wang, Sicheng and Zhang, Ruisi and Zhou, Meng and Zeng, Jiaqi and Dong, Xiangyu and Zhang, Ruoyu and others},
  booktitle={Proceedings of the 2020 conference on empirical methods in natural language processing (EMNLP)},
  pages={9241--9250},
  year={2020}
}

@inproceedings{shi2023midmed,
  title={MidMed: Towards mixed-type dialogues for medical consultation},
  author={Shi, Xiaoming and Liu, Zeming and Wang, Chuan and Leng, Haitao and Xue, Kui and Zhang, Xiaofan and Zhang, Shaoting},
  booktitle={Proceedings of the 61st Annual Meeting of the Association for Computational Linguistics (Volume 1: Long Papers)},
  pages={8145--8157},
  year={2023}
}

@inproceedings{johri2024craft,
  title={CRAFT-MD: A conversational evaluation framework for comprehensive assessment of clinical LLMs},
  author={Johri, Shreya and Jeong, Jaehwan and Tran, Benjamin A and Schlessinger, Daniel I and Wongvibulsin, Shannon and Cai, Zhuo Ran and Daneshjou, Roxana and Rajpurkar, Pranav},
  booktitle={AAAI 2024 Spring Symposium on Clinical Foundation Models},
  year={2024}
}

@misc{qiao2026medconsultbench,
      title={MedConsultBench: A Full-Cycle, Fine-Grained, Process-Aware Benchmark for Medical Consultation Agents}, 
      author={Chuhan Qiao and Jianghua Huang and Daxing Zhao and Ziding Liu and Yanjun Shen and Bing Cheng and Wei Lin and Kai Wu},
      year={2026},
      eprint={2601.12661},
      archivePrefix={arXiv},
      primaryClass={cs.AI},
      url={https://arxiv.org/abs/2601.12661}, 
}

@inproceedings{tang2016inquire,
  title={Inquire and diagnose: Neural symptom checking ensemble using deep reinforcement learning},
  author={Tang, Kai-Fu and Kao, Hao-Cheng and Chou, Chun-Nan and Chang, Edward Y},
  booktitle={NIPS workshop on deep reinforcement learning},
  year={2016}
}

@inproceedings{wei2018task,
  title={Task-oriented dialogue system for automatic diagnosis},
  author={Wei, Zhongyu and Liu, Qianlong and Peng, Baolin and Tou, Huaixiao and Chen, Ting and Huang, Xuan-Jing and Wong, Kam-Fai and Dai, Xiang},
  booktitle={Proceedings of the 56th Annual Meeting of the Association for Computational Linguistics (Volume 2: Short Papers)},
  pages={201--207},
  year={2018}
}

@inproceedings{xia2020generative,
  title={Generative adversarial regularized mutual information policy gradient framework for automatic diagnosis},
  author={Xia, Yuan and Zhou, Jingbo and Shi, Zhenhui and Lu, Chao and Huang, Haifeng},
  booktitle={Proceedings of the AAAI conference on artificial intelligence},
  volume={34},
  number={01},
  pages={1062--1069},
  year={2020}
}

@misc{grattafiori2024llama,
      title={The Llama 3 Herd of Models}, 
      author={Grattafiori, Aaron and Dubey, Abhimanyu and Jauhri, Abhinav and Pandey, Abhinav and Kadian, Abhishek and Al-Dahle, Ahmad and Letman, Aiesha and Mathur, Akhil and Schelten, Alan and Vaughan, Alex and others},
      year={2024},
      eprint={2407.21783},
      archivePrefix={arXiv},
      primaryClass={cs.AI},
      url={https://arxiv.org/abs/2407.21783}, 
}

@misc{openai2026gpt56,
  author       = {{OpenAI}},
  title        = {{GPT-5.6 System Card}},
  year         = {2026},
  howpublished = {OpenAI Deployment Safety Hub},
  url          = {https://deploymentsafety.openai.com/gpt-5-6},
  note         = {System card}
}

@misc{anthropic2026claudesonnet5,
  author       = {{Anthropic}},
  title        = {{Claude Sonnet 5 System Card}},
  year         = {2026},
  howpublished = {Anthropic},
  url          = {https://www.anthropic.com/claude-sonnet-5-system-card},
  note         = {System card}
}

@misc{googledeepmind2026gemini31pro,
  author       = {{Google DeepMind}},
  title        = {{Gemini 3.1 Pro Model Card}},
  year         = {2026},
  howpublished = {Google DeepMind},
  url          = {https://storage.googleapis.com/deepmind-media/Model-Cards/Gemini-3-1-Pro-Model-Card.pdf},
  note         = {Model card}
}

@misc{deepseek2026v4,
  author       = {{DeepSeek-AI}},
  title        = {{DeepSeek-V4: Towards Highly Efficient Million-Token Context Intelligence}},
  year         = {2026},
  howpublished = {Technical report},
  url          = {https://huggingface.co/deepseek-ai/DeepSeek-V4-Pro/blob/main/DeepSeek_V4.pdf}
}

@misc{lan2025clinicalgpt,
      title={ClinicalGPT-R1: Pushing reasoning capability of generalist disease diagnosis with large language model}, 
      author={Wuyang Lan and Wenzheng Wang and Changwei Ji and Guoxing Yang and Yongbo Zhang and Xiaohong Liu and Song Wu and Guangyu Wang},
      year={2025},
      eprint={2504.09421},
      archivePrefix={arXiv},
      primaryClass={cs.CL},
      url={https://arxiv.org/abs/2504.09421}, 
}

@misc{chen2024huatuo,
      title={HuatuoGPT-o1, Towards Medical Complex Reasoning with LLMs}, 
      author={Junying Chen and Zhenyang Cai and Ke Ji and Xidong Wang and Wanlong Liu and Rongsheng Wang and Jianye Hou and Benyou Wang},
      year={2024},
      eprint={2412.18925},
      archivePrefix={arXiv},
      primaryClass={cs.CL},
      url={https://arxiv.org/abs/2412.18925}, 
}

@misc{qwen3technicalreport,
      title={Qwen3 Technical Report}, 
      author={Qwen Team},
      year={2025},
      eprint={2505.09388},
      archivePrefix={arXiv},
      primaryClass={cs.CL},
      url={https://arxiv.org/abs/2505.09388}, 
}

@misc{yang2026mentalhospital,
      title={MentalHospital: A Virtual Environment for Evaluating Psychiatric Clinical Encounters}, 
      author={Yuming Yang and Xiao Sun and Yuanwei Zou and Zhengxiao Wu and Yun Chen and Jiang Zhong and Haoyang Zeng and Jingwang Huang and Kaiwen Wei},
      year={2026},
      eprint={2607.08257},
      archivePrefix={arXiv},
      primaryClass={cs.AI},
      url={https://arxiv.org/abs/2607.08257}, 
}

@misc{liang2026evaluating,
      title={Evaluating Large Language Models in Dynamic Clinical Decision-Making with Standardized Patient Cases}, 
      author={Cheng Liang and Pengcheng Qiu and Ya Zhang and Yanfeng Wang and Chaoyi Wu and Weidi Xie},
      year={2026},
      eprint={2606.05112},
      archivePrefix={arXiv},
      primaryClass={cs.CL},
      url={https://arxiv.org/abs/2606.05112}, 
}

@inproceedings{fan2025ai,
  title={Ai hospital: Benchmarking large language models in a multi-agent medical interaction simulator},
  author={Fan, Zhihao and Wei, Lai and Tang, Jialong and Chen, Wei and Siyuan, Wang and Wei, Zhongyu and Huang, Fei},
  booktitle={Proceedings of the 31st International Conference on Computational Linguistics},
  pages={10183--10213},
  year={2025}
}

@misc{bge-m3,
      title={BGE M3-Embedding: Multi-Lingual, Multi-Functionality, Multi-Granularity Text Embeddings Through Self-Knowledge Distillation}, 
      author={Jianlv Chen and Shitao Xiao and Peitian Zhang and Kun Luo and Defu Lian and Zheng Liu},
      year={2024},
      eprint={2402.03216},
      archivePrefix={arXiv},
      primaryClass={cs.CL}
}


\end{document}